# Deep vs. Diverse Architectures for Classification Problems

Colleen M. Farrelly[1]
[1] Kaplan University
`colleen.farrelly@kaplan.edu`

**Abstract**

This study compares various superlearner and deep learning architectures (machine-learning-based and neural-network-based) for classification problems across several simulated and industrial datasets to assess performance and computational efficiency, as both methods have nice theoretical convergence properties. Superlearner formulations outperform other methods at small to moderate sample sizes (500-2500) on nonlinear and mixed linear/nonlinear predictor relationship datasets, while deep neural networks perform well on linear predictor relationship datasets of all sizes. This suggests faster convergence of the superlearner compared to deep neural network architectures on many messy classification problems for real-world data.

Superlearners also yield interpretable models, allowing users to examine important signals in the data; in addition, they offer flexible formulation, where users can retain good performance with low-computational-cost base algorithms.

K-nearest-neighbor (KNN) regression demonstrates improvements using the superlearner framework, as well; KNN superlearners consistently outperform deep architectures and KNN regression, suggesting that superlearners may be better able to capture local and global geometric features through utilizing a variety of algorithms to probe the data space.

## 1 Introduction

Classification problems are ubiquitous in the medical and social sciences, and machine learning algorithms are increasingly used to solve these problems in criminology (Bogomolov et al., 2014; Kang & Choo, 2016; Pflueger et al., 2015; Stalans et al., 2004), psychology (Gowin et al., 2015; Karstoft et al., 2015), medicine (Aliper et al., 2016; Pirracchio et al., 2015; Tan & Gilbert, 2003;), and educational research (Blanch & Aluja, 2013; Dekker et al., 2009). Despite the success of machine learning in these fields, no single algorithm can perform optimally across problems, a result of The No Free Lunch Theorem (Fernandez-Delgado et al., 2014; Wolpert & Macready, 1997). Given this issue and the common challenges of outliers (Osberne & Overbay, 2004) and group overlap (Arabie, 1977; Huberty

& Lowman, 2000) in social science data, researchers must carefully choose which algorithm is likely to perform well on a given dataset (Fernandez-Delgado et al., 2014).

However, recent advances suggest two main families of algorithms do come with theoretical guarantees of performance and have found success in a variety of applications. One type of algorithm is the superlearner framework, a meta-learning ensemble algorithm comprised of multiple machine learning models (Van der Laan et al., 2007); asymptotically, this algorithm has a guarantee of performance at or above that of the best algorithm in the superlearner ensemble (Van der Laan et al., 2007) and has tackled difficult medical problems (Pirracchio et al., 2015). One advantage of this method is the potential diversity of algorithms utilized in the ensemble, a key factor related to classifier performance (Kuncheva & Whitaker, 2003). For example, an ensemble comprised of a main effects model (such as logistic regression), a classification tree model, and a kernel-based model can explore linear effects between a single predictor and outcome (logistic regression), more complex interactions representing subgroups (classification tree), and potential non-linear effects/space curvature (kernel-based model). Another advantage of the superlearner framework is the ability to assimilate knowledge from a variety of models while avoiding the issues related to multiple testing (Van der Laan et al., 2007); thus, the aforementioned example does not suffer from inflated discovery rates or introduced bias despite the use of multiple models.

A second type of framework with theoretical performance results includes feedforward neural networks. Many universal approximation theorems exist for feedforward neural networks with one hidden layer, suggesting that this architecture has a bounded error rate that approaches zero in the limit of the hidden layer size (Hornik et al., 1989; Huang et al., 2006; Huang & Wang, 2011); thus, an extremely wide feedforward neural network converges to perfect prediction. However, the width needed to attain arbitrarily small error may be computationally infeasible; deep learning offers one solution to this problem by deepening the neural network with additional hidden layers, creating a deep feedforward neural network that retains some of the theoretical properties of single-layer feedforward neural networks (Hartman et al., 1990). Deep learning has achieved widespread success on large industrial datasets and recent machine learning competitions (Schmidhuber, 2015); in addition, researchers and engineers have built important systems utilizing deep learning, including an emergency alert system (Kang & Choo, 2016).

One recent trend in deep learning has involved the mapping functions that transform input data from a previous layer to output data transmitted to the next layer (Zhou & Feng, 2017). Two noteworthy attempts include a random-forest-based deep learning architecture (Zhou & Feng, 2017) and a support-vector-based deep learning architecture (Tang, 2013). However, very little research exists on the use of machine learning algorithms as mapping functions within deep learning architectures, and it does not seem that any previous attempts have leveraged multiple machine learning methods within a single deep learning architecture. Because many machine learning algorithms converge at relatively small sample sizes (Breiman, 2001; Devroye, 1978; Lian, 2011), a deep learning framework utilizing these algorithms as mapping functions may be able to converge more quickly than those utilizing more traditional mapping functions, such as linear, sigmoid, or radial basis functions. In addition, these machine learning functions provide interpretable output to help with model diagnostics and answer data mining questions that may be relevant to understanding the classification problem.

This paper aims to compare superlearner ensembles with deep machine learning and neural network architectures to assess the potential for deep architectures blending multiple machine learning algorithms as mapping functions on a variety of simulated and real-world classification datasets. In addition, it aims to estimate relative gains above baseline algorithm performance by creating superlearner ensembles and deep architectures using the k-nearest-neighbor (KNN) regression model.

## 2 Methods

### 2.1 Superlearner and Deep Architecture Comparison: Simulations

To assess relative performance of algorithms, several frameworks competed against each other on a series of simulated datasets. The superlearner framework included random forest (Breiman, 2001), random ferns (Ozuysal et al., 2004), KNN regression (Altman, 1992), multivariate adaptive regression splines (MARS) (Friedman, 1991), conditional inference trees (Hothorn et al., 2006), and boosted regression with tree baselearners (Friedman & Meulman, 2003). Random forest, random ferns, boosted regression, and conditional inference trees provide for the identification of complex interaction terms and nonlinear relationships between predictors and outcome, while the MARS model captures linear relationships and main effects. KNN regression provides a way to capture local properties of the data, such as true outliers, dense clusters (large subgroups) that create curvature in the multivariate density function, and smaller subgroups. These techniques represent a variety of optimization strategies, including gradient-based, partitions based on entropy, and least squares solutions in a kernel space; this allows for adequate capture of dataset features (contours with gradients, clusters with entropy…) within a single ensemble. Algorithms were coded in R using ranger, randomFerns, caret, earth, ctree, and mboost package functions.

The deep feedforward machine learning model (mixed deep model) included 3 hidden layers. The first layer included two random forest models with different bootstrapping fractions, a conditional inference tree, and a random ferns model; the second layer consisted of a MARS model and another conditional inference tree, while the last layer included only boosted regression. This architecture was coded in R using the aforementioned packages, with data processing code at each layer.

Two deep neural network models were tested, one mirroring the mixed deep model framework (4-2-1 hidden layer sizes) and one tuned deep network with 4 hidden layers (13,5,3,1); both were constructed using the R package darch. Other parameters were tuned for optimality (including learning rate, number of epochs, batch size, bootstrap size, and momentum).

The KNN regression architectures included 1) a KNN regression model with k=5 and the kd_tree algorithm; 2) a deep architecture with 3 layers (10-10-5 configuration, same KNN regression model but bootstrap sampling within each layer to provide a variety of KNN regression estimates); and 3) a superlearner with varying k values (2, 5, 10, 25). Varying k allows for model diversity by considering finer and coarser geometric and topological features in the data, including dense clusters (creating curvature on the KNN graph), outliers, and small, locally optimal subgroups; these features often present algorithmic challenges within social science and medical datasets. Geometric features, such as local curvature of the data manifold, have played an important role in pattern-based recognition and matching (Li & Lu, 2017; Wang et al., 2012; Xu et al., 2010), as well as dimensionality reduction (Van Der Maaten et al., 2009; Wu & Wu, 2017) and network analysis (Saucan et al., 2017). Pairwise conservation of local properties in particular fosters improved performance of machine learning algorithms on real-world datasets (Van Der Maaten et al., 2009).

The simulations included a binary outcome (0, 1), 4 true predictors, and 9 noise predictors. Models were split by the nature of true predictor relationships with the outcome (linear only, nonlinear only, mixed) and Gaussian noise level (high, low). An additional condition of random misclassification/outliers representing group overlap between outcomes (5-10%) was added to the high noise level. This yielded 9 simulated datasets (linear—low noise, high noise, high noise + misclassification; nonlinear—low noise, high noise, high noise + misclassification; mixed—low noise, high noise, high noise + misclassification).

To examine the convergence properties of these algorithms within each simulation condition, 5 different sample sizes were generated for each condition with 500, 1000, 2500, 5000, and 10000 observations (10 replications each). A train/test split of 70/30 was used, and accuracy was assessed for each algorithm under each sample size and condition to compare performance (averaged across replications). All analyses were run on a 4-core i7/3GHz processor HP laptop with 16 gigabytes of memory.

## 2.2 Superlearner and Deep Architecture Performance: Real-World Data

To assess performance of optimally-tuned deep neural networks, machine-learning-based deep networks, and superlearners, algorithms were compared across 3 industrial datasets within Kaplan University studies. KNN frameworks were excluded, as performance was generally worse than the other methods on mixed, high noise, overlap conditions likely to exist in real-world data.

The first dataset aimed to predict Bar Exam passage rates for students attending Concord Law School in the past 10 years, including 188 students and 22 predictors that included admissions factors, law school performance, and performance on a standardized test required before moving from 1st Year to 2nd Year. Sampling used a 70/30 train/test split similar to the simulations.

The second dataset involved predicting retention among Kaplan University students attending online programs in late 2016 within a study assessing the usefulness of a new advising platform. This included 27,666 students and 10 predictors, including academic advising factors, student academic factors, and platform usage group. Again, train/test split employed 70/30 sampling.

The final dataset endeavored to predict student enrollment from admissions data. This included 905,612 leads within a 6-month period and 44 time-by-milestone factors. Because of a low enrollment fraction, stratified sampling was used to create a training set with a 20/80 split between training and test samples.

# 3 Results

## 3.1 Superlearner and Deep Architecture Comparison: Simulations

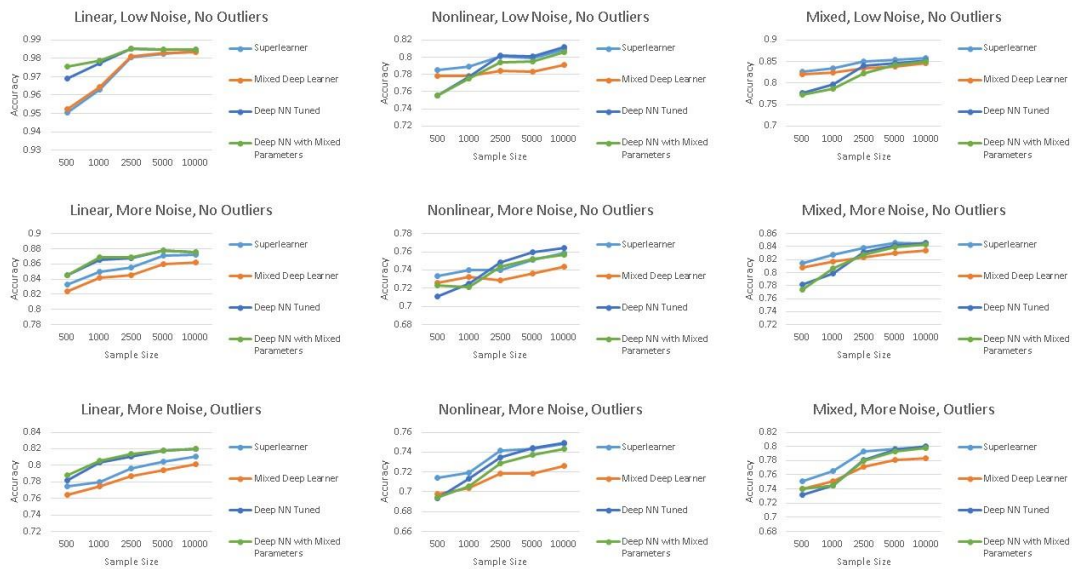

**Figure 1: Simulation comparison of deep learners vs. superlearners**

Simulation results suggest the efficacy of the superlearner framework for nonlinear or mixed predictor relationships, particularly at low sample sizes. Deep neural networks do eventually catch up to the superlearners, but this is mitigated by misclassification/outliers in the data. Machine-learning-based deep architectures show good performance relative to neural-network-based deep architectures at small sample sizes, but also struggle in the presence of in the presence of misclassification/outliers.

For linear relationships, deep neural networks perform well, even at small sample sizes. This suggests that they may be an optimal choice for problems involving many linear relationships, even if very little data exists upon which to train the algorithm. Because superlearners and machine-learning-based deep architectures utilized many tree-based algorithms, it is possible that performance was depressed by the use of methods typically used to capture nonlinear relationships between predictors and outcome.

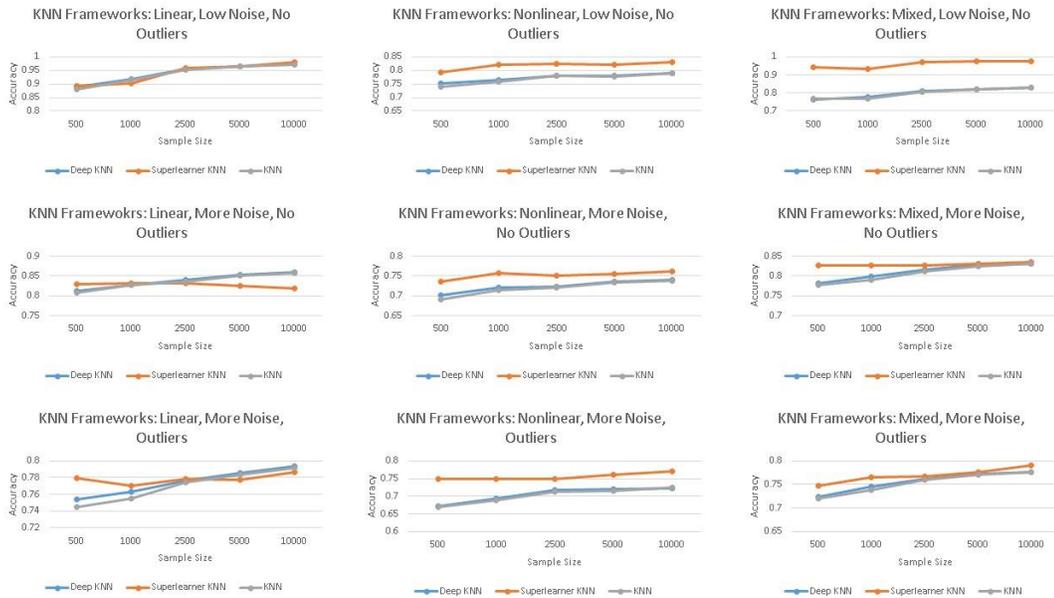

**Figure 2: KNN formulations for deep vs. superlearner gains**

The KNN formulations illustrate the efficacy of superlearner frameworks, particularly for small sample sizes and nonlinear/mixed relation predictors; in fact, most nonlinear/mixed relation trials show substantial gains over both the deep framework and the single KNN regression algorithm. This highlights the advantages of using a diverse learning ensemble for complex problems. Diverse ensembles seem to be able to capture more of the underlying function structure, likely by exploring more of the data space local geometry.

In addition, some gains exist from using the deep KNN architecture, particularly for noisy or small datasets. However, gains are not substantial, even with a highly-tuned framework.

## 3.2 Superlearner and Deep Architecture Comparison: Real-World Data

| Algorithm | Accuracy |
|---|---|
| Deep Machine Learning Network | 84.2% |
| Superlearner Model | 100.0% |
| Tuned Deep Neural Network | 68.4% |

**Table 1: Bar Exam passage classification task results**

The machine-learning-based superlearner and deep frameworks both outperform the tuned deep neural network substantially in predicting Bar Exam passage in a small sample of Concord Law Students; the superlearner achieves perfect performance on the test sample, suggesting the usefulness of this approach on small datasets with many nonlinear relationships.

The superlearner performance suggests the advantages of a diverse ensemble approach to exploring complex data spaces with few observations. The relative gain of the deep machine learning network relative to the deep neural network supports this, as well.

| Algorithm | Accuracy |
| --- | --- |
| Deep Machine Learning Network | 73.2% |
| Superlearner Model | 74.1% |
| Tuned Deep Neural Network | 74.4% |

**Table 2: Retention likelihood classification task results**

All algorithms perform equally well on predicting retention among students in terms selected for the advisor software trial, suggesting all algorithms have sufficiently converged. All algorithms also had similar runtimes of 2-5 minutes. Interestingly, the relationships discovered by the machine learning algorithms were mostly linear, mirroring the results of the linear, low noise trials.

| Algorithm | Accuracy | AUC | FNR | FPR | Time (Minutes) |
| --- | --- | --- | --- | --- | --- |
| Deep Machine Learning Network | 98.0% | 0.95 | 0.08 | 0.02 | 22 |
| Superlearner Model | 98.2% | 0.96 | 0.08 | 0.01 | 15 |
| Fast Superlearner Model | 98.0% | 0.95 | 0.08 | 0.02 | 2 |
| Tuned Deep Neural Network | 98.0% | 0.95 | 0.08 | 0.02 | 8 |

**Table 3: Admissions enrollment classification task results**

On the admissions classification task, all algorithms converge to similar accuracy and error rates, with the superlearner holding a slight advantage. The run-times for machine-learning-based frameworks were substantially longer than the deep neural network; however, a superlearner comprised of only a MARS model and a conditional inference tree model achieved similar accuracy with a much shorter runtime, suggesting that the flexibility offered by the superlearner framework can be used to optimize accuracy and runtime for larger datasets. This is a particular advantage with large industrial datasets. An additional simulation with a sample size of 10,000,000 for the mixed, misclassification/outliers trial was not computationally feasible for other algorithms on the laptop used, but ran and performed well using a superleaner comprised of conditional inference trees and linear models. For small companies and companies without extensive big data frameworks, being able to construct a robust algorithm that run quickly and perform well on large datasets using a local machine is key in deriving maximum benefit from analytics and accelerating discovery within industry.

# 4 Discussion

Across nonlinear and mixed simulations (and the industrial datasets), the superlearner frameworks showed earlier convergence properties and better performance on data with small sample sizes than deep architectures, particularly in the presence of misclassification/outliers. In addition, if presented with a large dataset, superlearners can include diverse algorithms that show good computational runtimes (such as conditional inference trees, MARS, and boosted linear models). Deep neural network models excelled at large sample sizes and purely linear relationships within the data simulations but showed poor performance at lower sample sizes, particularly with linear and nonlinear effects. The machine-learning-based deep architecture (mixed deep learner) performed well relative to deep neural network models at small sample sizes, particularly when there was no misclassification/group overlap/outliers.

This suggests that using machine learning algorithms used as mapping functions within a deep learning framework may provide a good solution when the size of training examples is small and offers the advantage of interpretable models, which may be of interest in data mining tasks. However, superelearners typically outperform these deep frameworks and show more robustness to misclassification/outliers/group overlap; they also enjoy the advantage of providing interpretable models.

Additional evidence for the efficacy of superlearners relative to deep frameworks can be seen in the KNN regression results. Although both frameworks provide some improvement over the single KNN regression model, the superlearner outperforms both models significantly in the presence of misclassification/outliers, as well as nonlinear or mixed relationships between predictors and the outcome. Diversity may be driving these results, as the superlearner explores different levels of features within the data, from very large (k=25) to very small (k=2) neighborhoods within the data space. This likely captures more of the geometry than any single model or any feedforward model based on optimal k-values.

Limitations of the study include examination of only one mixed deep learner framework; other formulations and optimal tuning may help improve its performance on simulated datasets and real-world problems. In addition, only one type of deep learning framework was used, and other deep networks, such as convolutional neural networks or deep belief networks, may show better performance on small datasets.

One interesting next step is to investigate the theoretical properties of the superlearner and determine if this framework could be formulated as a single-layer feedforward network with machine learning algorithms as mapping functions. This type of framework may enjoy universal approximation at smaller hidden layer sizes than traditional neural networks, analogous to the role of random mapping in extreme learning machines (ELMs). Random mapping within ELMs gives rise to universal approximation properties of ELMs (Huang et al., 2006), and deep learning has seen improvements in recent years by leveraging the properties of ELMs within a deep framework (Ding et al., 2015; Tang et al., 2016). Perhaps random mappings of machine learning functions can converge to universal approximation faster than ELMs and be leveraged for single-layer feedforward networks and deep architectures based on ELMs.